\documentclass[11pt,a4paper]{article}
\usepackage[hyperref]{acl2021}
\usepackage{times}
\usepackage{latexsym}

\usepackage{microtype}

\usepackage{graphicx}
\usepackage{subfigure}

\usepackage{amsmath}
\usepackage{bbding}

\aclfinalcopy 
\def\aclpaperid{1703} 

\title{Interventional Aspect-Based Sentiment Analysis}

\author{
Zhen Bi\textsuperscript{\rm 1,2 \thanks{\quad Equal contribution and shared co-first authorship.} }, 
Ningyu Zhang\textsuperscript{\rm 1,2 \footnotemark[1], \footnotemark[2]}, 
Ganqiang Ye\textsuperscript{\rm 1,2 \footnotemark[1]}, 
Haiyang Yu\textsuperscript{\rm 1,2 \footnotemark[1]}, 
Xi Chen\textsuperscript{\rm 3}, \\
\textbf{Huajun Chen}\textsuperscript{\rm 1,2 \thanks{\quad Corresponding author.}} \\

	\textsuperscript{\rm 1} Zhejiang University 
	\textsuperscript{\rm 2} AZFT Joint Lab for Knowledge Engine 
	\textsuperscript{\rm 3} Tencent \\
    \texttt{\{bizhen\_zju,zhangningyu,yeganqiang,yuhaiyang,huajunsir\}@zju.edu.cn} \\
	\texttt{jasonxchen@tencent.com} \\
}

\date{}

\begin{document}
\maketitle
\begin{abstract}
Recent neural-based aspect-based sentiment analysis approaches, though achieving promising improvement on benchmark datasets, have reported suffering from poor robustness when encountering confounder such as non-target aspects. In this paper, we take a causal view to addressing this issue. We propose a simple yet effective method, namely, Sentiment Adjustment (SENTA),  by applying a backdoor adjustment to disentangle those confounding factors. Experimental results on the Aspect Robustness Test Set (ARTS) dataset demonstrate that our approach improves the performance while maintaining accuracy in the original test set\footnote{The code and dataset are available in \url{https://github.com/zjunlp/SENTA}.}.
\end{abstract}

\section{Introduction}
Aspect-Based Sentiment Analysis (ABSA) is the task of classifying the sentiment polarity (positive, negative, neutral) on an aspect from a sentence or extracting aspects that reviewers have made comments on \cite{hu2004mining}.
Recently neural models have dominated the ABSA task, including memory networks \cite{wang2018target,tang2016aspect}, convolution methods \cite{li2018transformation,huang2019parameterized}, attention mechanism \cite{ma2017interactive} and dependency trees \cite{bai2020investigating}. 

However, open issues remain as neural models lack robustness for ABSA since they are sensitive to only the sentiment words of the target aspect, and therefore not be interfered with by the sentiment of any non-target aspect \cite{DBLP:conf/emnlp/XingJJWZH20}. 
For example, when there are multiple aspects in a review sentence, such as  ``The \textbf{pizza} is \textit{good} and \textbf{waiters} are \textit{friendly}.".
Current superior performance models usually suffers a accuracy decline in predicting the polarity of aspect \textbf{pizza} if the sentence is changed to ``The \textbf{pizza} is \textit{good} and \textbf{waiters} are \textit{unfriendly}.". 
The key challenge behind this phenomenon is caused by spurious correlations of statistical learning \cite{DBLP:conf/emnlp/ZengLZZ20}. 
From a causal perspective, spurious correlations are caused by confounding factors such as those other aspects in the same sentences. 
Based on the structural causal model (SCM) theory \cite{DBLP:journals/cacm/Pearl19}, if we intervene on the precursor variable in spurious correlations, we can eliminate those spurious correlations to some degree. 

\begin{figure}[tb]
\centering
\includegraphics[width=5cm]{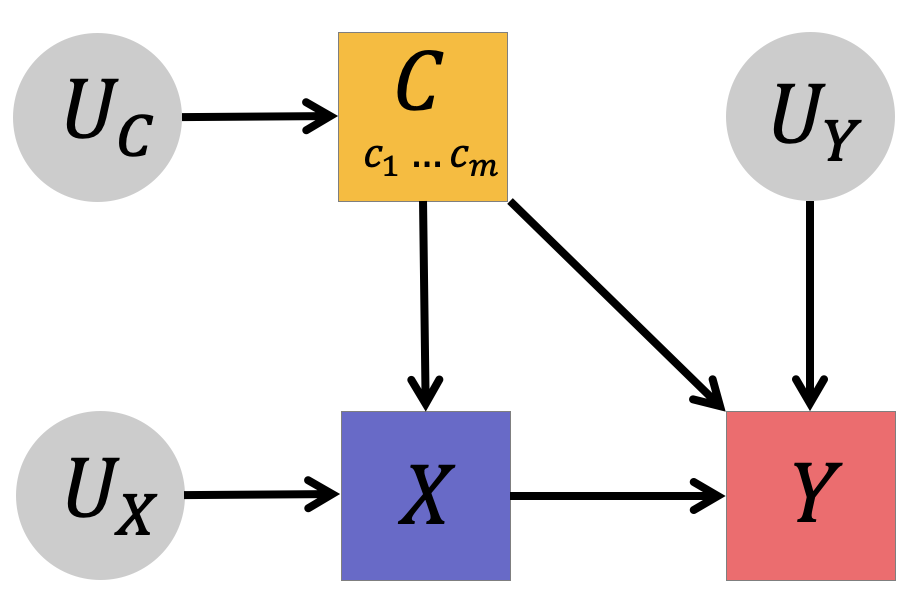}
\caption{The causal graph of ABSA. We build our causal model over three main variables:  \textbf{target feature} \( X \), \textbf{predictions}  \( Y \) and \textbf{confounding factor} \( C \). Our goal is to alleviate confounding factors, which is caused by \( X \gets C \), \( Y \gets C \).  }
\label{scm_} 
\end{figure}

Motivated by this, we propose the \textbf{SEN}timent \textbf{A}djustment (\textbf{SENTA}), which intervene between confounding factor and target aspect for ABSA. Firstly, we rethink the ABSA in the causal view in \S~\ref{step1} and introduce backdoor adjustment \cite{DBLP:journals/ai/Halpern19} in SCM, which try to intervene between confounding factor and target aspect in ABSA as shown in Figure \ref{scm_}.
Secondly, we introduce our Sentiment Adjustment approach in \S~\ref{step2}.
We train a confounding model without prior knowledge achieving good performance on the training and original test data. 
Then, we optimize a combination model to alleviate confounding effects by using decomposed confounding features. 
We evaluate our model's effectiveness in Aspect Robustness Test Set (ARTS), and our proposed method exhibits good performance compared with baselines.
Our major contributions are summarized as:
\begin{itemize}
\item We make the first attempt to take the causal view of ABSA to address the confounding factors. 
\item We propose a simple causal framework, Sentiment Adjustment, for ABSA, which obtain better performance than baselines. 
\end{itemize}

\section{Related Work}
ABSA  has recently emerged as an active research area with lots of approaches \cite{ma2017interactive,li2018transformation,huang2019parameterized,bai2020investigating}, yet challenges remain for robustness. 
\citet{DBLP:conf/emnlp/XingJJWZH20} introduce a new benchmark ARTS and probe the aspect robustness of neural models, and reveal up to \textbf{69.73\%} performance drop compared with the original test set. Previous work leverage  re-weighting \cite{DBLP:journals/corr/abs-1911-01460} to address this issue. Differently, we take the causal view of ABSA. 
Note that, causal inference  has been applied to various fields, including semantic segmentation \cite{DBLP:conf/nips/ZhangZT0S20}, few-shot learning \cite{DBLP:conf/nips/YueZS020}, etc. 
However, there are only a few works for natural language processing (NLP). \cite{DBLP:journals/corr/abs-2010-12919} propose an estimator and proves bias is bounded when performing an adjustment for the text. \cite{DBLP:journals/corr/abs-2012-04698} introduce a framework to generate counterfactual samples in text generation.
To the best of our knowledge, we are the first to apply causal inference to ABSA.

\section{Methodology}
In ABSA task, given a aspect $ t=(t_1, t_2, ..., t_m) $ about a product and a review sentence $s=(s_1, s_2, ..., s_n)$ containing the information about $t$. 
Aspect $t$ appears as a text span in sentence $s$ and a sentence may contain more than one aspect. 
The goal is to find polar sentiment (positive, neutral, negative) about specific aspect $t$.

\subsection{ABSA in the Causal View}
\label{step1}
Causal relations describe the causal effect among variables, which exist as the edge between nodes in SCM. Such relations are written using the assignment operator \( \gets \) and deterministic function notation \( f \), labeling the variable they affect. For example, we use  \( X \gets u_{x} \) represent the causal relationship of an unobserved variable on variable $X$. All causal relations in SCM is a directed acyclic graph (DAG). As shown in Figure \ref{scm_}, the SCM presented in the paper can be shown as follows:

 \[ X \gets f_x(C, U_{X}) \]
 \[ Y \gets f_y(X, C, U_{Y}) \]
 \[ C \gets f_c(U_{C}) \]
 
We build our causal model over three observed variables \textbf{target feature} \( X \) , \textbf{predictions}  \( Y \) and \textbf{confounding factor} \( C \). Variable \( U \) is called extraneous or unobserved variable, and \( u_{c} \) , \( u_{x} \) and \( u_{y} \) independent and unobserved noise variables.  As confounding factor \( C \) has impacts on \( X\),  we can get \( X \gets f_x(C, U_{X}) \). Variable \( C \) also has causal impact on predictions \( Y \), so
conditional distribution \( P (Y \mid X, Y, U) \) can be converted into \( Y \gets f_y(X, C, U_{Y}) \). To find the inner causal connection between \( X\) and \( Y\), we need to eliminate the influence of confounding factor \( C \).

\begin{figure*}[t]
    \centering
    \includegraphics[width=1\textwidth]{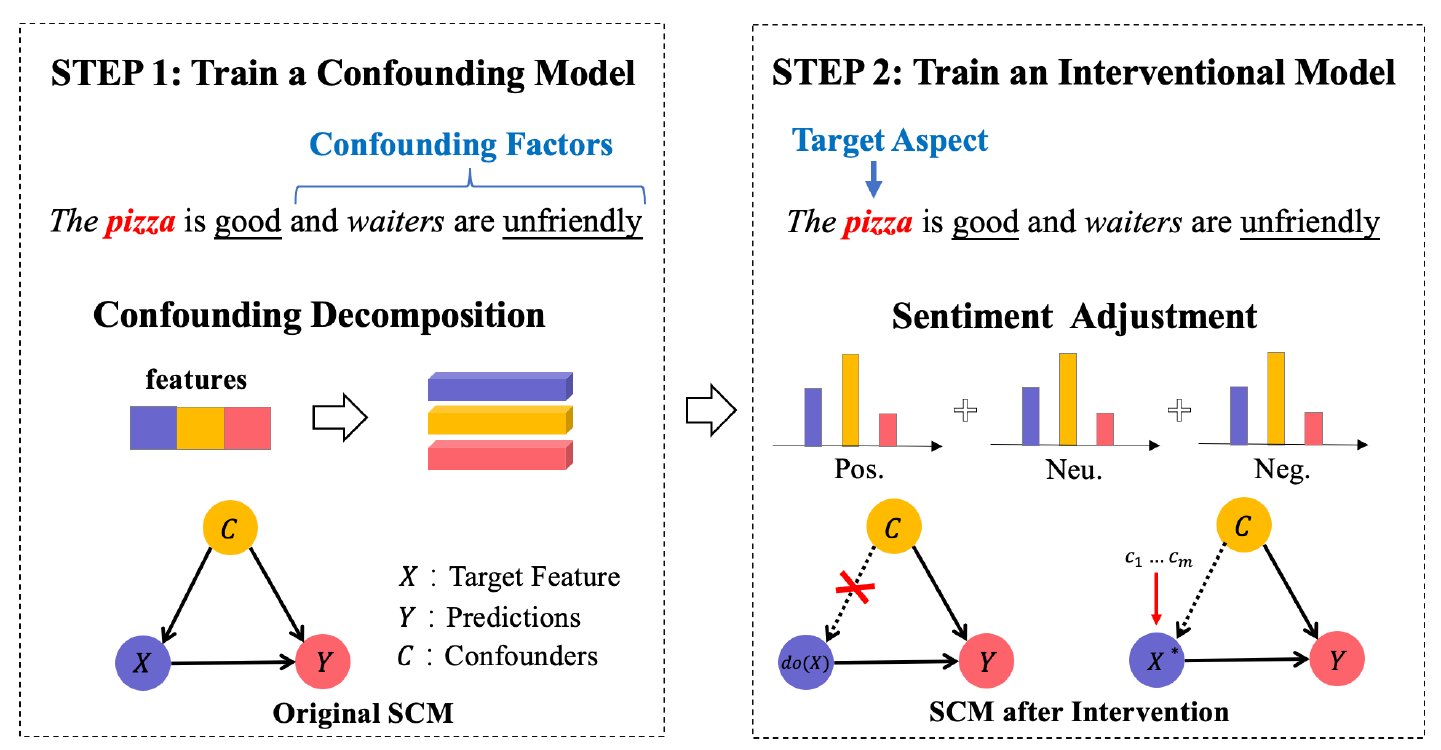}
    \caption{Framework of \textbf{SEN}timent \textbf{A}djustment (\textbf{SENTA}).}
    \label{model}
\end{figure*}

\subsection*{Backdoor Adjustment}
To intervene in SCM, \( do \) \( operation \) is used to describe the whole process. We use \( do(X=x) \) to express the intervention. When we do intervention to make \( X=x \), this process is denoted as \( P (Y=y \mid do(X=x) ) \).
If there are a set of variables \( C \) that satisfies the backdoor criterion (Appendix A), we can estimate the causal effect of $X$ on $Y$. As confounding factor meets the requirement, to know the effect of \( X \) (target feature) on \( Y \) (predictions), we regard variable \( C \) as the control, then make backdoor adjustment:

\[ P (Y=y \mid do(X=x) ) \]
\[ = \sum_{c} P (Y=y \mid X=x, C=c ) P(C=c) \]

Suppose there are \( m \) classes in classification,
then:

\[ = \sum_{i=1}^{m} P (Y=y \mid X=x, c_m ) P(c_m) \]
\[ \Rightarrow P\left(Y \mid x \oplus \frac{1}{m} \sum_{i=1}^{m} P\left(c_{i} \mid x\right) \bar{x_i}\right) \]

Note that \( P(c_{i} \mid x )\), which we denote as the output of corresponding class.
The key point is that, we make adjustment to original input $x$ by adding the decomposed class-level features of the trained confounding model.  

\subsection{Sentiment Adjustment}
\label{step2}
\subsection*{STEP1: Training a Confounding Model}

Since ABSA is to classify the sentiment of a specific aspect from a review sentence, it is similar to QA tasks. However, the output of ABSA is the polarity of aspect instead of a text span.
We leverage BERT to encode the input as \( ([CLS], q_1, ..., q_{N}, \) \( [SEP],s_1,...,s_{M}, [SEP]) \). Then we apply the hidden representation of \( h_{[CLS]} \) to the linear transformation to predict the sentiment polarity. 
The first step aims to obtain a model with good performance on the original test set but significantly deteriorates the new test set's performance.

\subsection*{STEP2: Training an Interventional Model}

As shown in Figure \ref{model},  we build an interventional framework. 
$h(\bar{x_i})$ is the mean hidden feature of the confounding model from class $c_i$.
\( h_{x_i} \) is the hidden states of main model.

There are \( m \) classifying polarities \( C=\{c_1, ... ,c_m \} \), 
given a training sample $\{x, y\}$, then

\[ \alpha_i = f_{classifer}(h(x_i))  \]
\[ h_C = \sum_{i}^{} \alpha_i h(\bar{x_i}) \]
\[ h_{adjust} =  f_{concat}(h_{M}, h_{C}) \]


\begin{table}
\centering
\begin{tabular}{ccccc}
\hline
    & \multicolumn{2}{c}{\textbf{Laptop}} & \multicolumn{2}{c}{\textbf{Restaurant}} \\
    & Ori              & Change              & Ori            & Change           \\
\hline
Positive    & 341          & 883        & 728     & 1,953          \\   
Negative    & 128          & 587        & 196     & 1,104          \\
Neutral     & 169          & 407        & 196     & 473           \\
\hline
\end{tabular}
\caption{Statistics of test sets}
\label{dataset_statistics}
\end{table}

\begin{table*}
\centering
\begin{tabular}{lllll}
\hline
\textbf{Dataset} & \multicolumn{2}{l}{\textbf{Laptop}} & \multicolumn{2}{l}{\textbf{Restaurant}} \\
\hline
Test    & Ori              & Change              & Ori            & Change           \\
\hline
BERT            & 75.07 & 63.71({\color{red} $\downarrow$11.36 }) & 82.50 & 73.37({\color{red} $\downarrow$9.13 }) \\   
BERT-Distill    & 75.54 & 65.64({\color{red} $\downarrow$9.90 }) & 81.61 & 71.53({\color{red} $\downarrow$10.08 }) \\
BERT-SENTA        & 75.08 & \textbf{67.23}({\color{red} $\downarrow$\textbf{7.85} }) & 83.30 & \textbf{77.30}({\color{red} $\downarrow$\textbf{6.00} }) \\
\hline
BERT-PT         & 80.25 & 71.82({\color{red} $\downarrow$8.43 }) & 86.60 & \textbf{80.99} ({\color{red} $\downarrow$5.61 }) \\ 
BERT-PT-Distill & 79.62 & 66.17({\color{red} $\downarrow$13.45 }) & 85.71 & 80.82({\color{red} $\downarrow$\textbf{4.89} }) \\
BERT-PT-SENTA    & 80.88 & \textbf{74.16}({\color{red} $\downarrow$\textbf{6.72} }) & 86.34 & 80.91({\color{red} $\downarrow$5.43 }) \\
\hline
\end{tabular}
\caption{
The accuracy of each model on the original test set and the new test set (ARTS) in laptop and restaurant domains. \textit{Ori} is the original test set in SemEval-2014 and \textit{Change} is ARTS.
}
\label{main_result}
\end{table*}

\section{Experiments}

\subsection{Datasets and Settings}

For evaluating our SENTA model, we use \emph{SemEval-2014 Task 4}\footnote{\url{https://alt.qcri.org/semeval2014/task4/index.php?id=data-and-tools}} in both laptop and restaurant domains for training, which is a popular benchmark for ABSA.
Specifically, we use the SemEval-2014 original (\textit{Ori}) test set as well as ARTS\footnote{\url{https://github.com/zhijing-jin/ARTS_TestSet}} (\textit{Change}) which is a  aspect robustness probing test set from \cite{DBLP:conf/emnlp/XingJJWZH20}. 
Statistics about test sets is shown in Table \ref{dataset_statistics}, \textit{Ori} is the original test set in SemEval-2014 and \textit{Change} is ARTS. 

\subsection{Baselines}
We compare with several baseline methods with the same hyper-parameters for fairness as follows: \textit{BERT} \cite{devlin2018bert} is \textit{BERT-base-uncased}, which is regarded as a baseline pretraining model in our experiment.
\textit{BERT-PT} \cite{xu2019bert} is a post-training language model, post-trained (fine-tuned) on a combination of Amazon reviews and all Yelp data.
\textit{BERT-PT} remains almost SOTA in ARTS so far and \textit{BERT(-PT)-Distill} \cite{hinton2015distilling} is a distillation method to combine confounding model with ABSA model. 
The training epochs for all models is set according to the evaluation in \textit{Ori}, instead of \textit{Change} which is unseen in a real scenario.  

\subsection{Results Analysis}
Results are shown in Table \ref{main_result}, including accuracy of six models on Laptop and Restaurant test sets as well as corresponding ARTS test set. Apparently, all methods perform worse in \textit{Change} than \textit{Ori}.
For example, in Laptop \textit{BERT} shows a sharp decline in \textit{Change} test set from \( 75.07 \% \) to  \( 63.71 \% \), and \textit{BERT-SENTA} declines from \( 75.08 \% \) to  \( 67.23 \% \). 
It shows that all methods still suffer from bias from confounding factors in new test set, which is hard to remove completely.  

\textit{SENTA} outperforms other methods while maintaining accuracy in the original test set. 
Since \textit{SENTA} is pluggable, we demonstrate its effectiveness with \textit{BERT} and \textit{BERT-PT} as the backbone, which improve the baseline model in \textit{Change} test set. 
It is  worth noting that post-training helps alleviate the confounding bias in \textit{Change} test set.

We  also list the declining accuracy (red numbers) of all methods in new test set. If \textit{BERT} is the baseline model, \textit{BERT-SENTA} has the least performance drop ( \( \downarrow 7.85 \% \) in Laptop and \( \downarrow 6.00\% \) in Restaurant) than others. 
If \textit{BERT-PT} is the baseline model, the falling range of \textit{BERT-PT-SENTA} is the least  ( \( \downarrow 6.72\%  \) in Laptop).
\textit{SENTA} shows weaker performances  ( \( 80.91\% \downarrow 5.43 \% \) in Restaurant), due to the effect of post-training. 

\subsection{Ablation Study}
\textbf{REVNON} \cite{DBLP:conf/emnlp/XingJJWZH20} is a strategy in generating ARTS, which could test whether a model is sensitive enough by perturbing the sentiments of the non-target aspects.
We split the ARTS and get REVNON subset results are shown in Table \ref{Ablation}. 
The more detailed case study is shown in Appendix B.

There are 444 and 135 REVNON's instances in Laptop and Restaurant domains. 
We compare SENTA with \textit{BERT} and \textit{BERT-PT}.  
Although we change the relative contents of non-target aspects, SENTA is still robust enough to bias from non-target aspects.
Our model performs better than other methods, further confirming the effectiveness of its mechanism.

\section{Conclusion and Future Work}

In this paper, we take the causal view of ABSA to address the robustness issue. 
We propose a novel Sentiment Adjustment (SENTA) model based on the backdoor adjustment to weaken confounding effects.
Experimental results demonstrate that our approach yields better performance on the robust set while maintaining accuracy in the original test set.
Our framework is general in the sense that any backbone models with different architectures can be employed.
In the future, we plan to  1) apply our approach to more NLP tasks with robustness issues and 2) find more reasonable metrics in evaluating the robustness of ABSA.

\begin{table}
\centering
\begin{tabular}{lcc}
\hline \textbf{Dataset} & \textbf{Laptop} & \textbf{Restaurant} \\ \hline
BERT  & 65.93 & 75.45 \\
BERT-SENTA  & \textbf{67.23} & \textbf{77.31}\\
BERT-PT  & 72.59 & 80.74 \\
BERT-PT-SENTA  & \textbf{74.16} & \textbf{80.91}\\
\hline
\end{tabular}
\caption{ Model accuracy in \textbf{REVNON} subset.
}
\label{Ablation}
\end{table}

\clearpage
\section*{Broad Impact Statement}

The causal inference has a wide range of applications, presenting researchers with an effective method to deeply understand relations between observed and unobserved variables.
Our work proves causal inference helps to analyze fine-grained sentiment classification task. 
Sentiment bias is a challenging and unsolved problem. 
Some social bias in sentiment analysis, including specific attributes (race, genders, occupations) is sensitive.
However, data-driven neutral models tend to fail in prediction because of bias from human annotation, bringing about unnecessary perplexity and trouble.
If researchers do not consider biases, it will be unfair for those with specific background or identification. 
Therefore we are supposed to provide the public with a qualified analyzing model in the application robust enough to harmful biases. 

\bibliographystyle{acl_natbib}
\bibliography{anthology, acl2021}

\appendix
\section{Backdoor Criterion}
If we want to know the effect of $X$ on $Y$ and have a set of variables $S$ as control, and $S$ satisfies the backdoor criterion if
\begin{itemize}\label{criterion}
\item \( S \) blocks every path from \( X \) to \( Y \) that has an arrow to \( X \).
\item  No node in \( C \) is a descendant of \( X \).
\end{itemize}

Then
\[ Pr (Y=y \mid do(X=x) ) \]
\[ = \sum_{s} Pr (Y=y \mid X=x, S=s ) Pr(S=s) \]

\section{Case Study}
\label{case_study}
We choose cases from \textbf{REVNON} in Laptop domain, results are shown in Table \ref{case1} and Table \ref{case2}.

\begin{table*}
\centering
\begin{tabular}{ccc}
\hline
CASE ID    & SENTENCE    &  POLARITY \\
\hline
1053:13\_0 & \multicolumn{1}{p{220pt}}{The \underline{\textbf{SD card reader}}  is slightly recessed but upside down (the nail slot on the card can be accessed), if this was not a self ejecting slot this would not be an issue, but its not.} & negative \\
\hline
1053:13\_1 & \multicolumn{1}{p{220pt}}{The SD card reader is slightly not recessed but not upside down (the nail \underline{\textbf{slot}} on the card can be accessed), if this was a self ejecting slot this would not be an issue, but its not.} & negative \\
\hline
1053:13\_2 & \multicolumn{1}{p{220pt}}{The SD card reader is slightly not recessed but not upside down (\underline{\textbf{the nail slot on the card}} cannot be accessed), if this was not a self ejecting slot this would not be an issue, but its not.} & negative \\ 
\hline
\end{tabular}
\caption{
The \textbf{REVNON} cases from Laptop domain. \underline{\textbf{Underlined}} words are target aspects.
}
\label{case1}
\end{table*}

\begin{table*}
\centering
\begin{tabular}{lcccc}
\hline
Method    & CASE 0 &  CASE 1 &  CASE 2\\
\hline
BERT & {\color{red}\XSolidBrush} & \color{red}\XSolidBrush & \color{green}\CheckmarkBold \\   
\hline
BERT-SENTA  & {\color{green}\CheckmarkBold}  & \color{green}\CheckmarkBold & {\color{green}\CheckmarkBold} \\
\hline
BERT-PT & {\color{red}\XSolidBrush} & \color{red}\XSolidBrush & \color{green}\CheckmarkBold \\ 
\hline
BERT-PT-SENTA  &  {\color{green}\CheckmarkBold}  & \color{green}\CheckmarkBold & {\color{green}\CheckmarkBold}  \\
\hline
\end{tabular}
\caption{
Comparison with different methods. Case statistics is shown in table \ref{case1}. {\color{green}\CheckmarkBold} denotes correct prediction and  {\color{red}\XSolidBrush} denotes wrong prediction. 
}
\label{case2}
\end{table*}

\section{Experiments Details}
We detail the training procedures and hyperparameters for each of the datasets. We utilize Pytorch to conduct experiments with one NVIDIA 1080 Ti 12GB GPU and support parallel training. All optimization was performed with the Adam optimizer. The max length for encoders is 64. More details can be seen in REDAME.md in the supplementary material.

\end{document}